\documentclass[nohyperref]{article}
\usepackage{microtype}
\usepackage{graphicx}
\usepackage{subfigure}
\usepackage{booktabs} 
\usepackage{hyperref}

\usepackage[accepted]{icml2022}

\usepackage{amsmath}
\usepackage{amssymb}
\usepackage{mathtools}
\usepackage{amsthm}
\usepackage[capitalize,noabbrev]{cleveref}
\theoremstyle{plain}

\theoremstyle{definition}

\theoremstyle{remark}

\usepackage{nicefrac}  
\usepackage{microtype}    
\usepackage{amsmath,amsfonts,bm}
\usepackage{algorithm,algorithmic}

\def\1{\bm{1}}

\def\vx{{\bm{x}}}
\def\vy{{\bm{y}}}
\def\vz{{\bm{z}}}

\DeclareMathAlphabet{\mathsfit}{\encodingdefault}{\sfdefault}{m}{sl}
\SetMathAlphabet{\mathsfit}{bold}{\encodingdefault}{\sfdefault}{bx}{n}

\newcommand{\R}{\mathbb{R}}

\usepackage{amssymb}

\usepackage{xcolor}
\definecolor{mydarkblue}{rgb}{0,0.08,0.45}
\definecolor{mygreen}{rgb}{0.032, 0.6392, 0.2039}
\definecolor{mypurple}{HTML}{B266FF}

\usepackage[inline, shortlabels]{enumitem} 
\DeclarePairedDelimiter{\norm}{\lVert}{\rVert}
\DeclarePairedDelimiterX{\inner}[2]{\langle}{\rangle}{#1,#2}		
\icmltitlerunning{Continuous-time Analysis for Variational Inequalities: An Overview and Desiderata}

\begin{document}
\onecolumn
\icmltitle{Continuous-time Analysis for Variational Inequalities:\\
An Overview and Desiderata}

\icmlsetsymbol{equal}{*}

\begin{icmlauthorlist}
\icmlauthor{Tatjana Chavdarova}{equal,ucb}
\icmlauthor{Ya-Ping Hsieh}{equal,eth}
\icmlauthor{Michael I. Jordan}{equal,ucb}
\end{icmlauthorlist}

\icmlaffiliation{ucb}{UC Berkeley}
\icmlaffiliation{eth}{Department of Computer Science, ETH
Zurich}

\icmlcorrespondingauthor{Tatjana Chavdarova}{tatjana.chavdarova@berkeley.edu}
\icmlcorrespondingauthor{Ya-Ping Hsieh}{yaping.hsieh@inf.ethz.ch}
\icmlcorrespondingauthor{Michael I. Jordan}{jordan@cs.berkeley.edu}
\icmlkeywords{Variational Inequality, Continuous-time}

\vskip 0.3in
\printAffiliationsAndNotice{\icmlEqualContribution}

\begin{abstract}
Algorithms that solve zero-sum games, multi-objective agent objectives, or, more generally, variational inequality (VI) problems are notoriously unstable on general problems. 
Owing to the increasing need for solving such problems in machine learning, this instability has been highlighted in recent years as a significant research challenge. In this paper, we provide an overview of recent progress in the use of continuous-time perspectives in the analysis and design of methods targeting the broad VI problem class.  Our presentation draws parallels between single-objective problems and multi-objective problems, highlighting the challenges of the latter.
We also formulate various desiderata for algorithms that apply to general VIs and we argue that achieving these desiderata may profit from an understanding of the associated continuous-time dynamics. 
\end{abstract}

\section{Introduction}
\emph{Variational inequalities} (VIs) are a broad class of mathematical programming problems that go beyond single-objective optimization to include complementarity problems~\citep{cottle_dantzig1968complementary}, zero-sum games~\citep{vonneumann1947gametheory,rockafellar1970monotone} and multi-player games.  The general VI problem takes the following form~\citep{stampacchia1964formes}:
\begin{equation} \label{eq:vi} \tag{VI}
  \mbox{find}\ \vz^\star \in \mathcal{Z} \quad\text{s.t.}\quad \langle \vz-\vz^\star, F(\vz^\star) \rangle \geq 0, \quad \forall \vz \in \mathcal{Z} \,,
\end{equation}
where $F: \mathcal{Z} \mapsto \R^d$ is a continuous map.  We focus on the unconstrained case in which $\mathcal{Z}\equiv\R^d$. 
Consider for example two agents, $\vx \in 
\R^{d_1}$ and $\vy \in \R^{d_2}$, that share a loss/utility function, $f \colon \R^{d_1} \times \R^{d_2} \to \R$ with $d_1 + d_2 \equiv d$,
which the first agent aims to minimize and the second agent aims to maximize. 
Then the problem is to find a 
\textit{saddle point} of $f$, i.e., a point $(\vx^\star,\vy^\star)$ such that
$
   f(\vx^\star,\vy)\leq f(\vx^\star,\vy^\star) \leq f(\vx,\vy^\star).
$
This corresponds to a VI in which $F(\vz)\equiv
\begin{bmatrix}\nabla_{\vx} f(\vx, \vy) & -\nabla_{\vy} f(\vx, \vy) \end{bmatrix}^\intercal$, with $\vz \triangleq \begin{bmatrix}
\vx &  \vy \end{bmatrix}^\intercal $.

A key difference between general VIs and single-objective minimization is that the Jacobian matrix of the gradient map, which for the zero-sum game is defined as:
$
  J(\vz) \!=\! \begin{bmatrix}
              \nabla_{x}^2 f(\vz)           &  \nabla_{y} \nabla_{x} f(\vz) \\
             -\nabla_{x} \nabla_{y} f(\vz)  & -\nabla_{y}^2 f(\vz)
             \end{bmatrix},
$
is generally \textit{non-symmetric}, resulting in dynamics that may \textit{rotate} around a fixed point.
For example, in the simple bilinear game, $f(\vx,\vy)\equiv \vx^\intercal \vy$, where $f$ is convex in $\vx$ and concave in $\vy$, the last iterate of the simple gradient descent ascent (GDA) method oscillates around the solution for an infinitesimal step size and \textit{diverges away} from it otherwise. 
This behavior is undesirable for many applications of VIs and consequently, the problem of designing algorithms with \textit{last-iterate} convergence has attracted significant attention 
\citep{goodfellow2016nips,daskalakis2018training,mescheder2018training,daskalakis2018limit,mazumdar2018convergence,MertikopoulosPP18,adolphs2018local,chavdarova2019,golowich20, chavdarova2021hrdes}.
Various methods have been proposed to resolve this problem, including the extragradient method \citep[EG,][]{korpelevich1976extragradient}, optimistic gradient descent ascent \citep[OGDA,][]{popov1980}, Halpern iteration~\citep{halpern1967}, and the lookahead-minmax algorithm \citep[LA,][]{chavdarova2021lamm}. 

Despite the recent progress on specific problem classes and on methods that exploit specific structure of $f$~\citep[see, e.g., ][]{diakonikolas2021efficient}, the general understanding of min-max optimization remains very limited relative to single-objective optimization. In particular, while methods such as EG and OGDA converge when $F$ is monotone---that is, when
$
\langle \vz-\vz', F(\vz)-F(\vz') \rangle \geq 0, \forall \vz, \vz' \in \mathcal{Z}
$---these same methods do not necessarily converge for general problems.  Indeed, they may rapidly diverge away from equilibria~\citep[see, e.g.,][]{chavdarova2021lamm}, or may converge to a limit cycle in which the method revisits a set of non-solution points with probability $1$~\citep{hsieh2020limits}.
These empirical observations are in sharp contrast to that of minimization, for which the Morse-Sard theorem asserts that the set of critical values when $F$ is a gradient field has Lebesgue measure $0$~\citep{sard1942,morse1939}. 
On the theoretical front, \citet{hsieh2020limits} showed that there exists a large class of problems for which all the popular min-max methods almost surely get attracted by a spurious limit cycle.  Thus, new algorithmic ideas are needed.

In the context of single-objective optimization with first-order oracles, numerous valuable insights have been found via a continuous-time perspective, in which continuous-time limits are taken of discrete-time algorithms, yielding a system of ordinary differential equations (ODEs)--see e.g., ~\citep{polyak1964some,arrow1957,Helmke96optimizationand,schropp2000,SuBoydCandes2016,muehlebach2019}. 
A particularly useful aspect of these ODEs is that they provide powerful analytic tools, such as Lyapunov theory, which are cumbersome to develop in discrete time. Moreover, rates of convergence obtained from the continuous-time theory can often be translated back into discrete time with an analogous proof guided by the continuous-time analysis. This is true for both upper and lower bounds. 

The continuous-time approach has not yet proved fruitful in the case of saddle point problems, however, because the existing first-order methods (i.e., GDA, EG, OGDA, and LA) lead to \emph{exactly the same ODE} in the limit when a naive conversion is carried out by taking the step size to zero \citep{hsieh2020limits,chavdarova2021hrdes}.  Thus, the qualitative differences observed empirically for these methods cannot be captured via the naive continuous-time limit. 
Motivated by this observation, recent work has considered \emph{higher-precision} ODEs which are able to track more closely the original discrete-time methods.  This approach distinguishes among existing methods and it provides theoretical convergence guarantees for some classes of problems; we review this line of work in \S~\ref{sec:related_works}.

Another important question in multi-objective learning---left largely unexplored---is the \emph{stability} of the discrete-time algorithms \cite{KY97}, which pertains to the \emph{non-divergence} of iterative schemes. In stark contrast to minimization where stability is often an immediate consequence of the descent inequalities in (expected) function values or gradient norms, the case of multi-agent setting is more subtle. In \S~\ref{sec:desiderata}, we discuss how the continuous-time perspective offers a principled avenue to establish stability for \emph{any} system, single-agent or multi-agent. Combined with the high-precision ODE framework alluded to above, this ultimately leads to practical training algorithms with theoretical convergence guarantees.

In this paper, we primarily provide an overview of the main ideas and results for continuous-time modeling of the VI-related problems. 
In addition, we aim to elaborate the distinguishing aspects of incorporating continuous-time based analysis, as well as list some such open problems in the VI context.

\section{Continuous-time Analyses for VIs: An Overview of Ideas and Insights}\label{sec:related_works}
We first describe the derivation of a ``standard'' ODE from a discrete-time update rule.
There are two distinct approaches to finding a correspondence between a discrete-time optimization method and a continuous-time ODE, which we describe in this section. We also review recent contributions that exploit the interpretability of the continuous-time dynamics to achieve acceleration.

\noindent\textbf{The classical derivation of an ODE.}
Consider an update rule of the discrete-time method in the following general form: 
$$
  \frac{\vz_{n+1}-\vz_n}{\gamma} \!=\! \mathcal{U}(\vz_{n+k}, \dots, \vz_0) \,,
$$
where $\gamma$ denotes the step size, $n$ the current step, and the abstract term $\mathcal{U}$ denotes the update rule of the discrete algorithm; e.g., for GDA: $\mathcal{U}\!\equiv \! -F(\vz_n)$.
To derive the associated ODE, one introduces the ansatz $\vz_n \!\approx\! \vz(n\delta)$, where $\delta$ denotes the time-step.
Letting $\delta = \gamma \to 0$ for GDA, yields the classical dynamics:
\begin{equation}\tag{GDA-ODE} \label{eq:ode-gda}
  \dot{\vz}(t) = - F\big(\vz(t)\big) \,.
\end{equation}
While providing useful insight into GDA, unfortunately this limiting procedure yields the \emph{identical} ODE for EG, OGDA, and LA, thus failing to distinguish among these methods~\citep[assuming diminishing step sizes; see][for details]{hsieh2020limits}, and in particular fails to capture the qualitative differences observed for bilinear games. Hence, a more refined analysis is required to  differentiate among these algorithms.

\noindent\textbf{Alternative discretizations.}
Even in single-objective minimization, it is sometimes the case that different algorithms yield the same underlying ODE, and in these cases, it can be fruitful to attempt to derive different algorithms as different numerical discretizations of the ODE~\citep{guilherme2021}.
Indeed, in the context of single-objective optimization, the same ODE can arise from different methods when there is a difference in the acceleration terms. In the~\ref{eq:vi} context, however, such an approach has not yet proved fruitful, perhaps because the differences among algorithms are qualitative.  A more fruitful approach has been to aim for different ODEs from the outset, using more sophisticated limiting procedures, as we now discuss.

\noindent\textbf{Deriving high-precision ODEs.}
Recent work aims at deriving ``higher-precision'' ODEs that more closely model the behavior of a discrete-time saddle-point method.
~\citet{rosca2021} use backward error analysis to construct modified ODEs that better describe the trajectories taken by the discrete-time updates and~\citet{lu2021osrresolution} proposes a general hierarchical procedure to construct an ODE given a discrete-time method.
In recent work,~\citet{chavdarova2021hrdes} adapt a general methodology from fluid dynamics~\citep{fluidmechanicsbook}, called \textit{High-Resolution 
Differential Equations} (HRDEs), which have also been similarly employed in the analysis of minimization methods \citep{shi2018hrde}. 
This procedure generalizes the naive ODE derivation by constructing an $\mathcal{O}(\gamma^m)$-HRDE that retains all the terms that depend on the powers $\gamma^0,\dots, \gamma^m$ and discarding those terms that depend on $\gamma^{m+1}, \gamma^{m+2}, \dots$.  The resulting HRDE can be analyzed via various continuous-time tools.  For example, when the derived HRDE is linear, the authors show that the Routh–Hurwitz stability criterion~\citep{xie1985} can be usefully employed to study the convergence.  This classical technique from dynamical systems theory consists in the construction of a simple array of coefficients and based on their signs allows for determining if the system is stable, \textit{without} explicitly deriving the eigenvalues of the associated matrix of the system. A particularly noteworthy result is that by first rewriting the HRDE of OGDA in an equivalent form as in~\citep{attouch2012second}, the authors are able to recover the original OGDA method using simple Euler discretization.

\noindent\textbf{Nonlinear HRDEs.}
\citet[][Table 1]{chavdarova2021hrdes} derived HRDEs corresponding to the GDA, EG, and LA-GDA methods, showing that they have the following form:
\vspace{-.4em}
$$
\ddot{\vz} (t) +  \frac{2}{\gamma} \cdot \dot{\vz}(t) = - \alpha \cdot F\big( \vz(t) \big) + \beta  \cdot J\big( \vz(t) \big) F\big( \vz(t) \big)\,, \qquad \text{with:} \quad
\alpha,\beta \in \R_+ \,. \vspace{-.3em}
$$
The different methods have different $\alpha,\beta$ coefficients.
The authors showed that, when $F$ is a general vector field, the first term contributes to the rotational component, and indeed it is non-zero only when $F$ is rotational. The second term adds to the potential component since it influences the continuous-time dynamics to decrease the norm of the operator. The GDA method, which is known to diverge on monotone VI problems, has $\beta=0$ for GDA, whereas all the remaining methods have $\beta\neq0$. 
Also, the fact that the HRDEs of LA-GDA (with $k=2$) and the EG methods are identical reveals that these are related (up to an error term of $\gamma^3$), a fact that is not revealed by naive inspection of their update rule.
For the HRDE of OGDA, the second term is instead $J\big( \vz(t) \big) ) \dot{\vz}(t)$ and is analogous to the Hessian-driven damping of minimization methods~\citep{alvarez2002,attouch2019}.

\noindent\textbf{Deriving new methods by discretization of ODEs.}
\citet{hemmat2020LEAD} view the current iterate (in the parameter space) as a \textit{particle} in a dynamical system, while modeling also its rotational force and adding a compensating force to guarantee convergence on quadratic games. By discretizing the proposed ODE, the authors obtain a second-order update rule.
\citet{bot2020} study 
the asymptotic convergence of the continuous-time
trajectory generated by a forward-backward-forward dynamics and its explicit discretization reveals Tseng's forward-backward-forward algorithm~\citep{tseng2000}.
Inspired by the ODE of the accelerated methods in minimization,~\citet{bot2022fast} design a second-order ODE which differs from those in~\citet{chavdarova2021hrdes} it that it has an asymptotically vanishing damping term.  Discretizing this ODE yields a method that resembles OGDA.

\noindent\textbf{Other tools from continuous-time systems.}
Much of the progress that stems from continuous-time perspectives comes from the use of Lyapunov theory.
\citet{fiez2021local} establish local convergence guarantees of GDA when using timescale separation, by combining Lyapunov stability  and a guard map~\citep{Saydy1990GuardianMA}.
\citet{zhang2021unified} propose a standardized way of finding the parameters of a \textit{quadratic} Lyapunov function of a given \textit{first-order} saddle point optimizer in discrete time, using the theory of integral quadratic constraints, but can be applied solely to strongly-monotone operator $F$.
\citet{nagarajan_gradient_2017} use an approximated continuous dynamical system to study the local stability of GANs, and \citet{bailey2019hamiltionian} delineate a connection  between zero-sum games and Hamiltonian systems. \citet{ryu2019} also build on insights provided by the continuous-time ODE analyses.

\section{Desiderata for Future Work}\label{sec:desiderata}

\paragraph{Stability.}
The core idea of proving the stability of a system is usually to establish that the system ``has a tendency to stay somewhere.'' This is illustrated by considering the continuous-time dynamics of the form $\dot{\vz}(t) = F(\vz(t))$, which implies
\begin{equation}\label{eq:del-norm}\tag{S}
    \frac{\mathrm{d}}{\mathrm{d}t}{\left(\frac{1}{2} \norm{\vz(t)}^2 \right)} = \inner{\vz(t)}{\dot{\vz}(t)} = \inner{\vz(t)}{F(\vz(t))}.
\end{equation}
Thus, if $\inner{\vz}{F(\vz)} \leq 0$ for all $\vz$ such that $\norm{\vz} \geq R>0$, then it immediately follows that the system cannot explode since \eqref{eq:del-norm} forces $\norm{\vz}$ to decrease outside of a compact set. Notice that we have imposed essentially no additional condition in this line of reasoning; in particular, the argument applies equally well to single- or multi-agent learning. 
 
It is thus natural to ask if the condition $\inner{\vz}{F(\vz)} \leq 0$ can also be exploited to guarantee the stability of the \emph{discrete-time} schemes. This connection is partially addressed in \citet{hsieh2020limits}, which provides an affirmative answer to the above question in the low-precision ODE case (i.e., via reduction to GDA-ODE). These authors also show that the condition can be seen as a generalization of the classical notion of coercivity, a mainstay in the analysis of stability for variational inequalities \cite{BC17}.

It is our view that the full power of the continuous-time perspective on stability analysis is under-explored for variational inequalities. In particular, it is not known whether we can translate the stability from HRDEs, or, more generally, any sufficiently regular ODE, to their discretized counterparts. In this regard, we expect the simple arguments outlined above to continue to guide the stability analysis of iterative schemes that are well beyond the scope of existing results.

\paragraph{Stochasticity.}

In the context of optimization, the connection between stochastic gradient descent and stochastic differential equation (SDE) theory has led to several surprising findings such as convergence to the global minima or finite hitting time \cite{zhang2017hitting,raginsky2017non}. The prominent feature of these results is that \emph{no convexity is assumed}, and are thus applicable to the state-of-the-art decision systems based on deep learning.

Surprisingly, we are not aware of any such result in the multi-agent or variational inequality settings. In particular, this link is missing even for the elementary stochastic GDA, whose stationary distributions exist under fairly general conditions \cite{khasminskii2011stochastic}. Investigating the properties of such distributions, such as concentration around the equilibria or limit cycles, as well as the corresponding discrete-time analysis, therefore presents important open questions.

\section{Conclusions}

We have pointed out some of the advantages of a continuous-time perspective on variational inequalities, which include
\begin{enumerate*}[series = tobecont, itemjoin = \enspace, label=(\roman*)]
\item increased interpretability, as the separate terms of the continuous-time dynamics may have physical meaning that can increase our understanding of its performance, and
\item design tools, as the continuous-time formulation provides means to derive new classes of algorithms
using various discretization  schemes. 
\end{enumerate*}
Moreover, establishing convergence guarantees in continuous time is often simpler, and often aids  in proving the corresponding result of the discrete-time counterparts. 
We also overviewed cases where the continuous-time formulation provides a unifying framework to analyze closely related discrete-time methods and to apply existing stability analyses from dynamical systems.
Several questions remain open in the context of general VIs, and we allude to the under-explored possibility of tackling broader classes of potentially non-monotone problems via continuous-time approaches.

\section*{Acknowledgements}
We acknowledge support from the Swiss National Science Foundation (SNSF), grant P2ELP2\_199740, from the Mathematical Data Science program of the Office of Naval Research under grant number N00014-18-1-2764, as well as from the ETH Foundations of Data Science (ETH-FDS) postdoctoral fellowship.

\bibliography{main}
\bibliographystyle{icml2022}
\end{document}